\documentclass[a4paper, 10pt, conference]{ieeeconf}    
                                                      
\usepackage{FG2018}

\usepackage{times}
\usepackage{epsfig}
\usepackage{relsize}
\usepackage{subfigure}
\usepackage{graphicx}
\usepackage{amsmath}
\usepackage{amssymb}
\usepackage{multirow}
\usepackage{gensymb}
\usepackage{tabularx, booktabs}
\newcolumntype{Y}{>{\arraybackslash}X}
\usepackage{pifont}
\usepackage{caption}

\FGfinalcopy

\def\mbf#1{\mathbf{#1}}

\def\tbf#1{\textbf{#1}}

\newcommand{\minisection}[1]{\vspace{2mm}\noindent{\bf #1}.}

\usepackage[pagebackref=true,breaklinks=true,letterpaper=true,colorlinks,bookmarks=false]{hyperref}

\IEEEoverridecommandlockouts                              
                                                         
\def\FGPaperID{172} 

\title{\LARGE \bf
ExpNet: Landmark-Free, Deep, 3D Facial Expressions
}

\author{Feng-Ju Chang\textsuperscript{1}, Anh Tuan Tran\textsuperscript{1}, Tal Hassner\textsuperscript{2,3}, Iacopo Masi\textsuperscript{1}, Ram Nevatia\textsuperscript{1}, Gerard Medioni\textsuperscript{1}\\
\textsuperscript{1}~Institute for Robotics and Intelligent Systems, USC, CA, USA\\
\textsuperscript{2}~Information Sciences Institute, USC, CA, USA\\
\textsuperscript{3}~The Open University of Israel, Israel\\
{\tt\small \{fengjuch,anhttran,iacopoma,nevatia,medioni\}@usc.edu, hassner@openu.ac.il}
}

\begin{document}

\IEEEoverridecommandlockouts\pubid{\makebox[\columnwidth]{978-1-5386-2335-0/18/\$31.00~\copyright{}2018 IEEE \hfill}
\hspace{\columnsep}\makebox[\columnwidth]{ }}

\ifFGfinal
\thispagestyle{empty}
\pagestyle{empty}
\else
\author{Anonymous FG 2018 submission\\ Paper ID \FGPaperID \\}
\pagestyle{plain}
\fi
\maketitle

\begin{abstract}
We describe a deep learning based method for estimating 3D facial expression coefficients. Unlike previous work, our process does not relay on facial landmark detection methods as a proxy step. Recent methods have shown that a CNN can be trained to regress accurate and discriminative 3D morphable model (3DMM) representations, directly from image intensities. By foregoing facial landmark detection, these methods were able to estimate shapes for occluded faces appearing in unprecedented in-the-wild viewing conditions. We build on those methods by showing that facial expressions can also be estimated by a robust, deep, landmark-free approach. Our ExpNet CNN is applied directly to the intensities of a face image and regresses a 29D vector of 3D expression coefficients. We propose a unique method for collecting data to train this network, leveraging on the robustness of deep networks to training label noise. We further offer a novel means of evaluating the accuracy of estimated expression coefficients: by measuring how well they capture facial emotions on the CK+ and EmotiW-17 emotion recognition benchmarks. We show that our ExpNet produces expression coefficients which better discriminate between facial emotions than those obtained using state of the art, facial landmark detection techniques. Moreover, this advantage grows as image scales drop, demonstrating that our ExpNet is more robust to scale changes than landmark detection methods. Finally, at the same level of accuracy, our ExpNet is orders of magnitude faster than its alternatives. 
\end{abstract}

\section{Introduction}\label{sec:intro}

Successful methods for single view 3D face shape modeling were proposed nearly two decades ago~\cite{blanz2002face,blanz2003face,paysan09basel,romdhani2003efficient}. These methods, and the many that followed, often claimed high fidelity reconstructions and offered parameterizations for facial expressions besides the underlying 3D facial shape. 

 \begin{figure}[tb]
 \centering
 \includegraphics[width=\linewidth]{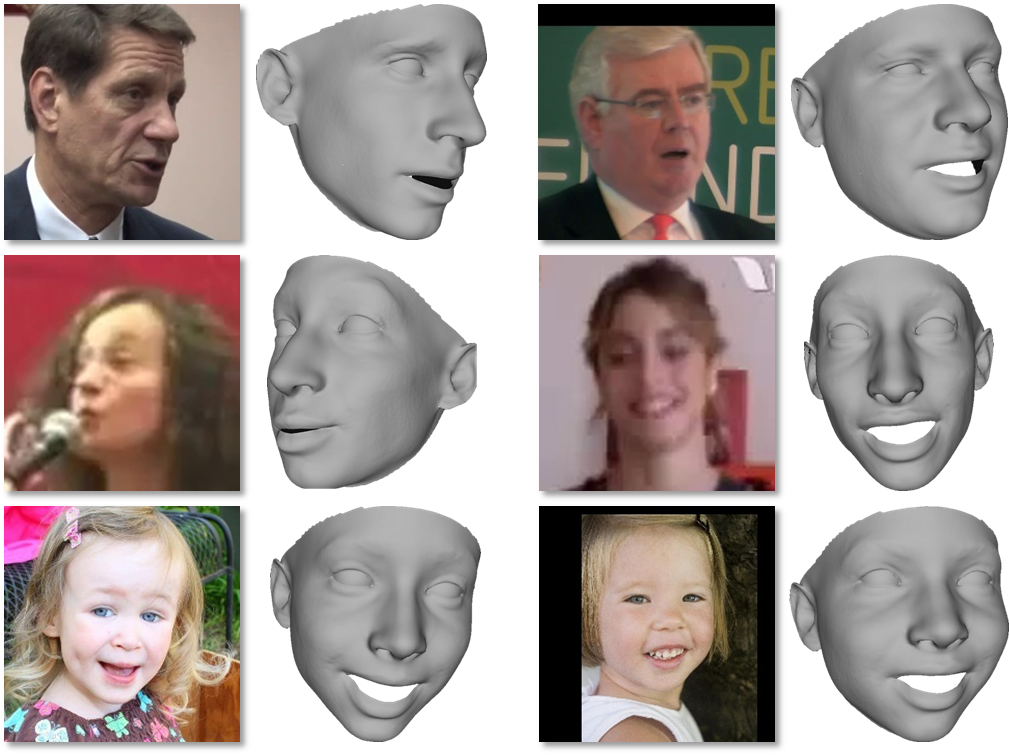}
 \caption{
 {\em Deep 3D face modeling with expressions}. We propose to regress 3DMM expression coefficients without facial landmark detection, directly from image intensities. We show this approach to be highly robust to extreme appearance variations, including out-of-plane head rotations (top row), scale changes (middle), and even ages (bottom).
 }
 \label{fig:teaser}
 \end{figure}

Despite their impressive results, they and others since~\cite{blanz2002face,blanz2003face,chu2014,paysan09basel,romdhani2003efficient,tang2008real,yang2011expression} suffered from prevailing problems when it came to processing face images taken under unconstrained viewing conditions. Many of these methods relied to some extent on facial landmark detection, performed either prior to reconstruction or concurrently, as part of the reconstruction process. By involving landmark detection, these methods are sensitive to face pose and, aside from a few recent exceptions (e.g., 3DDFA~\cite{zhu2015}), could not operate well on faces viewed in extreme out of plane rotations (e.g., near profile). Scale and occlusions are also problems: Whether because landmarks are too small to accurately localize or altogether invisible due to occlusions, accurate detection and consequent 3D reconstruction is not handled well. 

In addition to these problems, many methods applied iterative steps of {\em analysis-by-synthesis}~\cite{Bas:accvw16,huber:3dmm,romdhani2005estimating}. These methods were not only computationally expensive, but also hard to distribute and run in parallel on dedicated hardware such as the ubiquitous graphical processing units (GPU).

Very recently, some of these problems were addressed by two papers, which are both relevant to this work. First, Tran et al.~\cite{tran16_3dmm_cnn} proposed to use a deep CNN to estimate the 3D shape and texture of faces appearing in unconstrained images. Their CNN regressed 3D morphable face model (3DMM) parameters directly. To test the extent to which their estimates were robust and discriminative, they then used these 3DMM parameters as face representations in challenging, unconstrained face recognition benchmarks, including the Labeled Faces in the Wild (LFW)~\cite{LFWTech} and the IARPA Janus Benchmark A (IJB-A)~\cite{Klare_2015_CVPR}. By doing so, they showed that their estimated 3DMM parameters were nearly as discriminative as opaque deep features extracted by deep networks trained specifically for recognition.

Chang et al.~\cite{chang17fpn} extended this work by showing that 6 degrees of freedom (6DoF) pose can also be estimated using a similar deep, landmark free approach. Their proposed FacePoseNet (FPN) essentially performed face alignment in 3D, directly from image intensities and without the need for facial landmarks which are usually used for these purposes. 

Our paper uses similar techniques to model 3D facial expressions. Specifically, we show how facial expressions can be modeled directly from image intensities using our proposed deep neural network: ExpNet. To our knowledge, this is the first time that a CNN is shown to estimate 3D expression coefficients directly, without requiring or involving facial landmark detection.

We provide a multitude of face reconstruction examples, visualizing our estimated expressions on faces appearing in challenging unconstrained conditions (see, e.g., Fig.~\ref{fig:teaser}). We know of few previous method who offered this many examples of their capabilities. 

We go beyond previous work, however, by additionally offering {\em quantitative} comparisons of our facial expression estimates. To this end, we propose to measure how well different expression regression methods capture facial emotions on the Extended Cohn-Kanade (CK+)~\cite{lucey2010extended} and EmotiW-17 benchmarks~\cite{dhall2017individual}. Both benchmarks contain face images labeled for emotion categories, allowing us to focus on how well emotions are captured by our method and others. We show that not only does our deep approach provide more meaningful expression representations, it is more robust to scale changes than methods which rely on landmarks for this purpose. Finally, to promote reproduction of our results, {\em our code and deep models are publicly available.}\footnote{Available:~\url{github.com/fengju514/Expression-Net}}

\section{Related Work}\label{sec:relatedwork}
\subsection{Expression Estimation}
We first emphasize the distinction between the related, yet different tasks of {\em emotion classification} vs. {\em expression regression}. The former seeks to classify images or videos into discrete sets of facial emotion classes~\cite{dhall2017individual,lucey2010extended} or action unites~\cite{fabian2016emotionet,zafeiriou2016facial}. This problem was often addressed by considering the locations of facial landmarks. In recent years a growing number of state of the art methods have instead adopted deep networks~\cite{kosti2017emotion,levi2015emotion,zhang2016gender}, applying them directly to image intensities rather than estimating landmark positions as a proxy step.

Methods for expression regression attempt to extract parameters for face deformations. These parameters are often expressed in the form of active appearance models (AAM)~\cite{lucey2010extended} and Blendshape model coefficients~\cite{richardson2016learning,zhu2015,zhu2015high}. In this work we focus on estimating 3D expression coefficients, using the same representation described by 3DDFA~\cite{zhu2015}. Unlike 3DDFA, however, we completely decouple expression coefficient regression from facial landmark detection. Our tests demonstrate that by doing so, we obtain a method which is more robust to changing image scales.

\subsection{Facial Landmark Detection} There has been a great deal of work dedicated to accurately detecting facial landmarks, and not only due to their role in expression estimation. Face landmark detection is a general problem which has applications in numerous face related systems. Landmark detectors are very often used to align face images by applying rigid~\cite{eidinger2013age,Everingham06a,wolf:YTF} and non-rigid transformations~\cite{hassner2013viewing,jeni2015dense,zhu2015} transformations in 2D and 3D~\cite{hassner2015effective,Masi:18:learning,masi2014pose,masi2017rapid,masi16dowe}. 

Generally speaking, landmark detectors can be divided into two broad categories: {\em Regression based}~\cite{burgos2013robust,king2009dlib,wu2017facial} and {\em Model based}~\cite{baltruvsaitis2016openface,zadeh2016deep,zhu2015} techniques. Regression based methods estimate landmark locations directly from facial appearance while model based methods explicitly model both the shape and appearance of landmarks. Regardless of the approach, landmark estimation can fail whenever faces are viewed in extreme out-of-plane rotations (far from frontal), low scale, or when the face bounding box differs significantly from the one used to develop the landmark detector.

To address the problem of varying 3D poses, the recent 3DDFA~\cite{zhu2015}, related to our own, learns the parameters of a 3DMM representations using a CNN. Unlike us, however, they prescribe an iterative, analysis-by-synthesis approach. Also related to us is the recent CE-CLM~\cite{zadeh2016deep}. CE-CLM introduces a convolution expert network to capture very complex landmark appearance variations and thereby achieving state-of-the-art landmark detection accuracy.

The exact locations of facial landmarks were once considered subject-specific information which can be used for face recognition~\cite{edwards1998face}. Today, however, such attempts are all but abandoned. The reason for turning to other face representations may be due to the real-word imaging conditions typically assumed by modern face recognition systems~\cite{Klare_2015_CVPR} where even state of the art landmark detection accuracy is insufficient to discriminate between individuals based solely on the locations of their detected facial landmarks. In other applications, however, facial landmarks prevail. This work follows recent attempts, most notably Chang {\em et al.}~\cite{chang17fpn}, by proposing landmark free alternatives for face understanding tasks. This effort is intended to allow for accurate expression estimation on images which defy landmark detection techniques, in similar spirit to the abandonment of landmarks as a means for representing identities. To our knowledge, {\em such a direct, landmark free, deep approach to expression modeling was never previously attempted} .

\section{Deep, 3D Expression Modeling}\label{sec:deepexpr}
We propose to estimate facial expression coefficients using a CNN applied directly to image intensities. A chief concern when training such deep networks is the availability of labeled training data. For our purposes, training labels are 29D real-valued vectors of expression coefficients. These labels do not have a natural interpretations that can easily be used by human operators to manually collect and label training data. We next explain how 3D shapes and their expressions are represented and how ample data may be collected to effectively train a deep network for our purpose.

\subsection{Representing 3D Faces and Expressions}\label{sec:background}
We assume a standard 3DMM face representation~\cite{blanz2002face,blanz2003face,chu2014,hu2016face,paysan09basel}. Given an input face photo $\mbf{I}$, standard methods for estimating its 3DMM representation typically detect facial feature points and then use those as constraints when estimating the optimal 3DMM expression coefficients (see, for example, the recent 3DDFA method~\cite{zhu2015}). Instead, we propose to estimate expression parameters by directly regressing 3DMM expression coefficients, decoupling shape and texture from pose and from expression.

Specifically, we model a 3D face shape using the following, standard, linear 3DMM representation (for now, ignoring parameters representing facial texture and 6DoF pose):
\begin{equation}
\mbf{S}^{\prime} = \widehat{\mbf{s}} + \mbf{S} \boldsymbol{\alpha} + \mbf{E} \boldsymbol{\eta}
\label{eq:3DMM}
\end{equation}
where $\widehat{\mbf{s}}$ represents the average 3D face shape. The second term provides shape variations as a linear combination of shape coefficients $\boldsymbol{\alpha} \in \mathbb{R}^s$ with $\mbf{S}\in \mathbb{R}^{3n\times s}$ principal components. 3D expression deformations are provided as an additional linear combination of expression coefficients $\boldsymbol{\eta}\in \mathbb{R}^{m}$ and expression components $\mbf{E}\in \mathbb{R}^{3n\times m}$. Here, $3n$ represents the 3D coordinates for the $n$ pixels in $\mbf{I}$. The numbers of components, $s$, for shape and for expression, $m$, provide the dimensionality of the 3DMM coefficients. Our representation uses the BFM 3DMM shape components~\cite{paysan09basel}, where $s=99$ and the expression components defined by 3DDFA~\cite{zhu2015}, with $m=29$.

The vectors $\boldsymbol{\alpha}$ and $\boldsymbol{\eta}$ control the intensity of deformations provided by the principal components. Given estimates for $\boldsymbol{\alpha}$ and $\boldsymbol{\eta}$, it is therefore possible to reconstruct the 3D face shape of the face appearing in the input image using Eq.~(\ref{eq:3DMM}). 

\subsection{Generating 3D Expression Data}\label{sec:gen_data}
To our knowledge, there is no publicly available data set containing sufficiently many face images labeled with their 29D expression coefficients. Presumably, one way of mitigating this problem is to use a 3D facial expressions database such as BU-4DFE~\cite{yin2008high} as a training data set. BU-4DFE faces, however, are viewed under constrained conditions and this would therefore limit application of the network to constrained settings. Furthermore, BU-4DFE contains only 101 subjects and six facial expressions and can thus limit the range of expression coefficients our network predicts.

Another way of addressing the training data problem is by utilizing a face landmark detection benchmark. That is, taking the face images in existing landmark detection benchmarks and computing their expression coefficients using their ground truth landmark annotations in order to obtain 29D ground truth expression labels. Existing landmark detection benchmarks, however, are limited in their sizes: The number of images in the training and testing splits of the popular 300W landmark detection data set~\cite{sagonas2015300}, for example, is 3,026. This is far too small to train a deep CNN to regress 29D real valued vectors.

Given the absence of sufficiently large and rich 3D expression training sets, we propose a simple method for generating ample examples of faces in thew wild, coupled with 29D expression coefficients labels. We begin by estimating 99D 3DMM coefficients for the 0.5 million face images in the CASIA WebFace collection~\cite{yi2014learning}. 3DMM shape parameters were estimated following the state of the art method of~\cite{tran16_3dmm_cnn}, giving us, for every CASIA image, an estimate of its shape coefficients, $\boldsymbol{\alpha}$. We assume that all images belonging to the same subject should have the same, single 3D shape. We therefore apply the shape coefficients pooling method of~\cite{tran16_3dmm_cnn} to average the 3DMM shape estimates for all images belonging to the same subject, thereby obtaining a single 3DMM shape estimate per subject. 

Poses were additionally estimated for each image using FPN~\cite{chang17fpn}. We then use standard techniques~\cite{hartley2003multiple} to compute a projection matrix $\boldsymbol{\Pi}$ from the 6DoF provided by that method.

Given a projection matrix $\boldsymbol{\Pi}$ that maps from the recovered 3D shape, $\mbf{S}^{\prime}$, to the 2D points of an input image, we can solve the following optimization problem to get expression coefficients:
\begin{equation}
\begin{split}
\boldsymbol{\eta}^{\star} = \arg\min_{\boldsymbol{\eta}} || \mbf{p} - \boldsymbol{\Pi} \mbf{S}^{\prime} ||_2, \\
\text{subject to} \quad |\boldsymbol{\eta}_j| \leq 3~{\delta}_{\mbf{E}_j},
\end{split}
\label{eq:expr_fitting}
\end{equation}
where $\boldsymbol{\alpha}$ in $\mbf{S}^{\prime}$ (Eq.~\ref{eq:3DMM}) is estimated by~\cite{tran16_3dmm_cnn}. ${\delta}_{\mbf{E}_j}$ is the standard deviation of the j-th principal components of the 3DMM expression; $\mbf{p}$ is a set of 2D facial landmarks detected in the input image by a standard facial landmark detection method, in our experiments, CLNF~\cite{baltrusaitis2013constrained}. We solve for $\boldsymbol{\eta}^{\star}$ in Eq.~(\ref{eq:expr_fitting}) via standard Gauss-Newton optimization.

\subsection{Training ExpNet to Predict Expression Coefficients}\label{sec:training}
We use the expression coefficients obtained from Eq.~(\ref{eq:expr_fitting}) as ground truth labels when training our ExpNet. In practice, ExpNet employs a ResNet-101 deep network architecture~\cite{He_2016_CVPR}. We did not experiment with smaller network structures, and so a more compact network may well work just as well for our purposes. Our ExpNet is trained to regress a parametric function $f(\{\mbf{W},\mbf{b}\}, \mbf{I}) \mapsto \boldsymbol{\eta}$, where $\{\mbf{W},\mbf{b}\}$ represent the parametric filters and weights of the CNN. We use a standard $\ell_2$ reconstruction loss between ExpNet predictions and its expression coefficients training labels.

ExpNet is trained using Stochastic Gradient Descent (SGD) with a mini-batch size of~144, momentum of~0.9, and weight decay of~5e-4. The network weights are updated with learning rate set to~1e-3. When the validation loss saturates, we decrease learning rates by an order of magnitude until the validation loss stops decreasing. No data augmentation is performed during training: that is, we use the plain images in the CASIA set since they are already roughly aligned~\cite{yi2014learning}. In order to make training easier, we removed the empirical mean from all the input faces.

We note that our approach is similar to the one used by Tran et al.~\cite{tran16_3dmm_cnn}, and in particular, we use the same network architecture used in their work to regress 3DMM shape and texture parameters. They, however, explicitly assume a unique shape representations for all images of the same subject. This assumption allowed them to better regularize their network, by presenting it with multiple images with varying nuisance but the same underlying label (i.e. shape coefficients do not vary within the images of a subject). Here, this is not the case and expression parameters vary from one image to the next, regardless of subject identity.

\subsection{Estimating Expressions Coefficients with ExpNet}\label{sec:predict}
Existing methods for expression estimation often take an analysis-by-synthesis approach to optimizing facial landmark locations. Contrary to them, our expressions are obtained in a single forward pass of our CNN. To estimate an expression coefficients vector, $\boldsymbol{\eta}_t$, we evaluate $\mbf{I}_t$ $f(\{\mbf{W},\mbf{b}\}, \mbf{I}_t)$ for test image, $\mbf{I}_t$. We preprocess test images using the face detector of Yang et al.~\cite{yang2016multi} and increasing its returned face bounding box by a scale of $\times 1.25$ of its size. This scaling was manually determined to bring face bounding boxes to roughly the same size as the loose bounding boxes of CASIA faces.

\section{Experimental Results}\label{sec:expt}

\begin{figure*}[tb]
\centering
\subfigure[]{
\includegraphics[width=.3\linewidth]{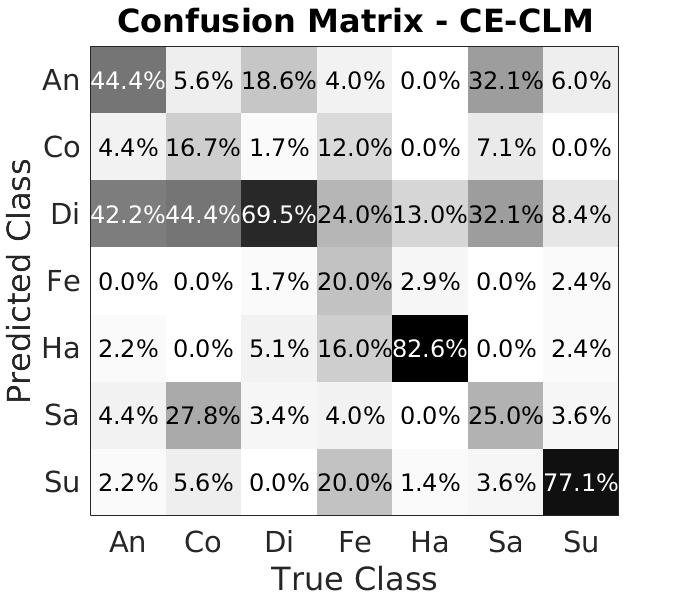}
\label{fig:exp_rec_conf_dclm}
}
~
\subfigure[]{
\includegraphics[width=.3\linewidth]{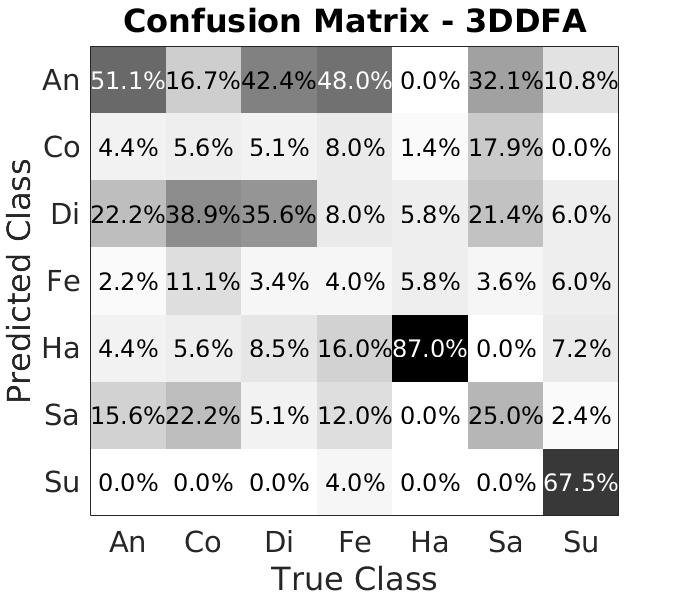}
\label{fig:exp_rec_conf_3ddfa}
}
~
\subfigure[]{
\includegraphics[width=.3\linewidth]{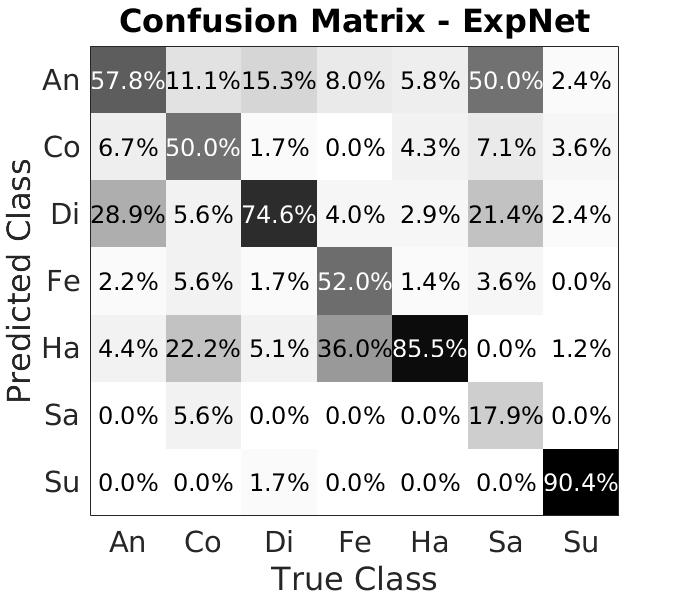}
\label{fig:exp_rec_conf_us}
}
\caption{
{\em Confusion matrix for emotion recognition on the CK+ benchmark~\cite{lucey2010extended}}. Confusion distributions across emotion classes using the original input image resolution. Results provided for (a) the best performing landmark detector, CE-CLM~\cite{zadeh2016deep}, (b) the recent, deep 3DDFA~\cite{zhu2015}, (c) our ExpNet.
}
\label{fig:exp_conf}
\end{figure*}

\begin{figure*}[tb]
\centering
\subfigure[]{
\includegraphics[width=.3\linewidth]{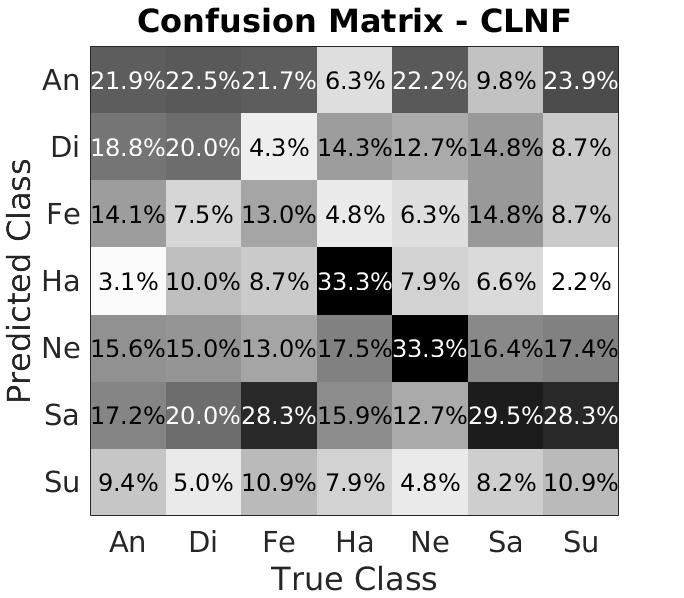}
\label{fig:exp_rec_conf_clnf_emotiw17}
}
~
\subfigure[]{
\includegraphics[width=.3\linewidth]{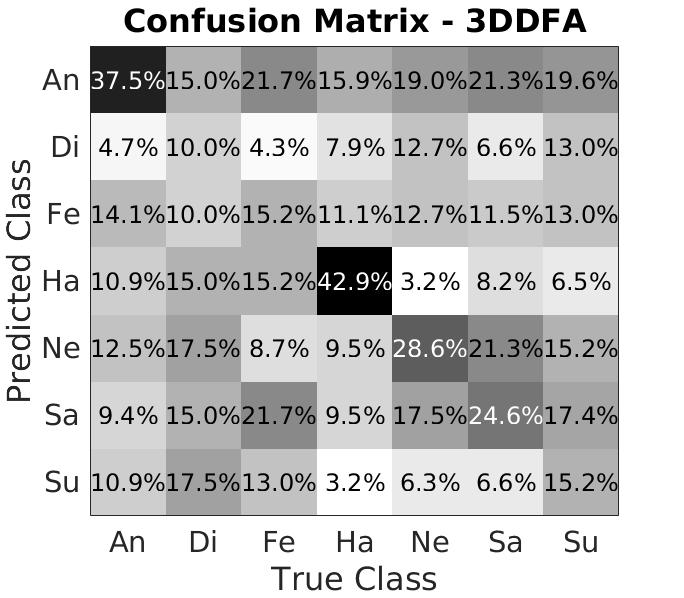}
\label{fig:exp_rec_conf_3ddfa_emotiw17}
}
~
\subfigure[]{
\includegraphics[width=.3\linewidth]{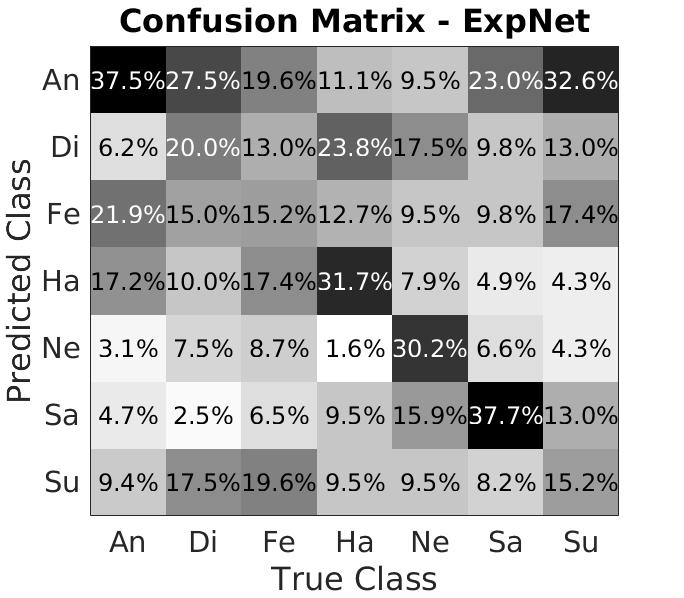}
\label{fig:exp_rec_conf_us_emotiw17}
}
\caption{
{\em Confusion matrix for emotion recognition on the EmotiW-17 benchmark~\cite{dhall2017individual}}. Confusion distributions across emotion classes using the original input image resolution. Results provided for (a) the best performing landmark detector, CLNF~\cite{baltrusaitis2013constrained}, (b) the recent, deep 3DDFA~\cite{zhu2015}, (c) our ExpNet.
}
\label{fig:exp_conf_emotiw17}
\end{figure*}

We evaluated our method both qualitatively and quantitatively. It is important to note that few previous methods for 3D expression estimation performed quantitative tests; instead, most offered only qualitative results. We provide an extensive number of figures demonstrating the quality of our expression estimation method (Sec.~\ref{sec:qual}) In addition, we offer quantitative tests, designed to capture the extent to which our expressions reflect facial emotions (Sec.~\ref{sec:quant}).  

\subsection{Quantitative Tests}\label{sec:quant}
\minisection{Benchmark settings} Aside from 3DDFA~\cite{zhu2015}, we know of no previous method which directly estimates 29D expression coefficients vectors. Instead, previous work relied on facial landmark detectors and used their detected landmarks to estimate facial expressions. We therefore compare the expressions estimated by our ExpNet to those obtained from state of the art landmark detectors. Because no benchmark exists with ground truth expression coefficients, we compare these methods on the related task of facial emotion classification. Our underlying assumption here is that better expression estimation implies better emotion classification.  

We use benchmarks containing face images labeled for discrete emotion classes. For each image we estimate its expression coefficients, either directly using our ExpNet and 3DDFA, or using detected landmarks by solving Eq.~(\ref{eq:expr_fitting}) as described in Sec.~\ref{sec:gen_data}. We then attempt to classify the emotions for test images using the exact same classification pipeline applied to these 29D expression representations. 

Our tests utilize the Extended Cohn-Kanade (CK+) dataset~\cite{lucey2010extended} and the Emotion Recognition in the Wild Challenge (EmotiW-17) dataset~\cite{dhall2017individual}. The CK+ dataset is a constrained set, with frontal images taken in the lab, while the EmotiW-17 dataset contains highly challening video frames collected from 54 movie DVDs~\cite{dhall2012collecting}.

The CK+ dataset contains 327 face video clips labeled for seven emotion classes:
anger (An), contempt (Co), disgust (Di), fear (Fe), happy (Ha), sadness (Sa), surprise (Su). From each clip, we take the peak frame (the end of video)– the frame assigned with an emotion label – and use it for classification. The EmotiW-17 dataset, on the other hand, offers 383 face video clips labeled for 7 emotion classes: anger (An), disgust (Di), fear (Fe), happy (Ha), neutral (Ne), sadness (Sa), surprise (Su).  We estimate 29D expression representations for every frame and apply average pooling of the per-frame estimates across all frames of each video. 

Following the protocol used by~\cite{lucey2010extended}, we ran a leave-one clip-out test protocol to assess performance. We also evaluate the robustness of different methods to scale changes. Specifically, we tested all methods on multiple version of the CK+ and EmotiW-17 benchmarks, each version with all images scaled down to $\times$0.8, 0.6, 0.4, and 0.2 their sizes.

\minisection{Emotion classification pipeline} The same simple classification method was used for all methods in all our tests. We preferred a simple classification method rather than a state of the art technique, in order to prevent obscuring the quality of the landmark detector / emotion estimation by using an elaborate classifier. We therefore use a simple kNN classifier with $K=5$. It is important to note that the results obtained by all of the tested methods are far from the state of the art on this set; our goal is not to outperform state of the art emotion classification methods, but only to compare expression coefficient estimation techniques. 

\begin{figure*}[tb]
\centering
\subfigure[]{
\includegraphics[width=.44\linewidth]{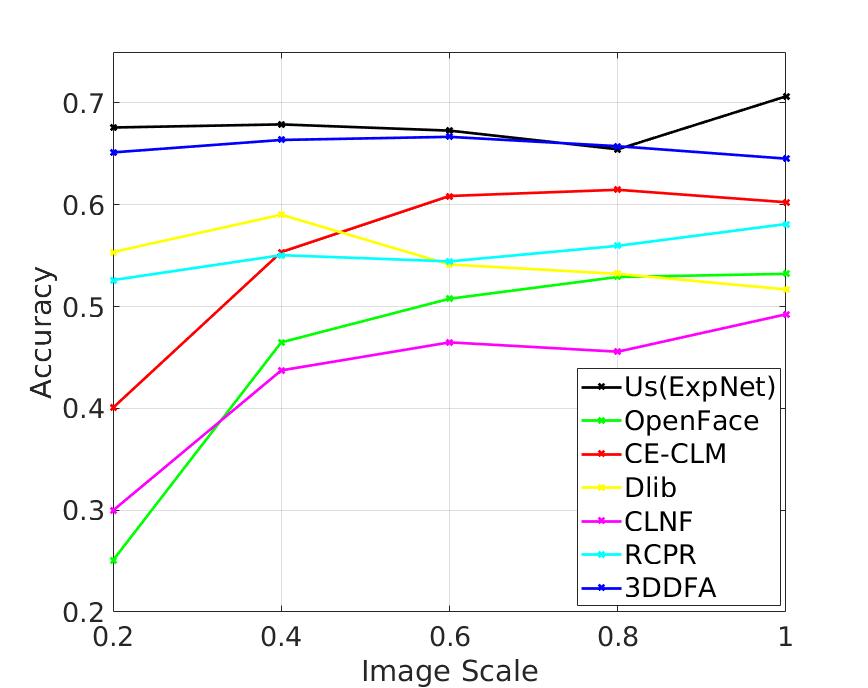}
\label{fig:exp_acc_ckplus}
}
~
\subfigure[]{
\includegraphics[width=.44\linewidth]{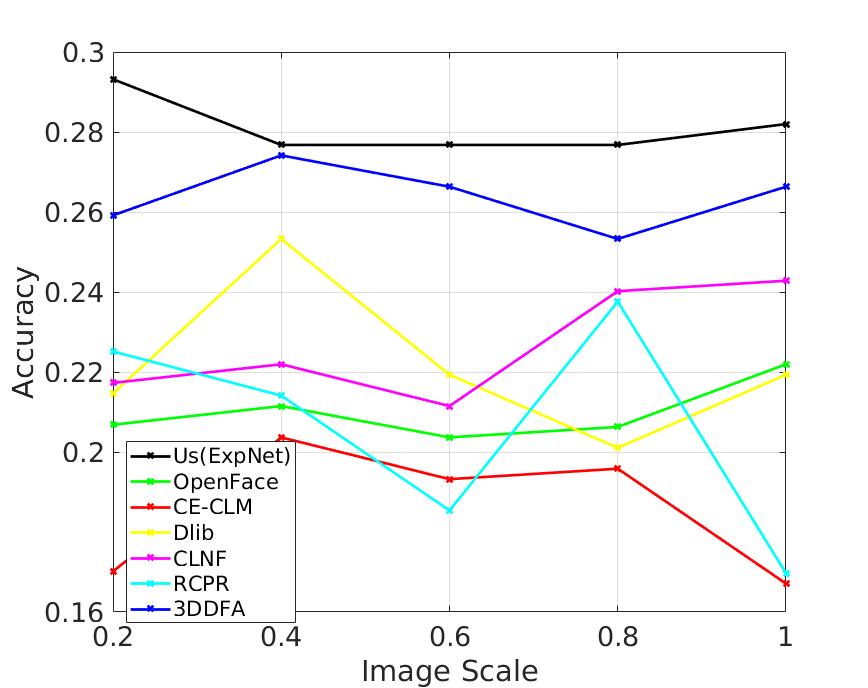}
\label{fig:exp_acc_emotiw17}
}
\caption{
{\em Emotion recognition accuracy across scales.} Results provided for (a) the CK+ and (b) EmotiW-17 benchmarks. Each curve corresponds to a different method. For each scale, the experiment resizes the input image accordingly. Lower scale values indicate lower resolutions. Original resolutions were 640$\times$490 (CK+) and 720$\times$576 (EmotiW-17).
}
\label{fig:exp_rec_scale}
\end{figure*}

\begin{table}[tb]
\Large
\centering
\resizebox{1.0\linewidth}{!}{
\begin{tabular}{l|ccccc|cc}
\toprule
			 	&	\multicolumn{5}{c|}{Landmark-based}	   			& \multicolumn{2}{c}{Deep, Direct}	\\[2ex] \hline
Time (s/img) 	& DLIB  & CE-CLM  & OpenFace & CLNF & RCPR  & 3DDFA & \tbf{Us (ExpNet)}  						\\[.4ex] \hline
Landmarks 		& 0.009	& 15.83	  &	0.31	 & 0.38	& 0.19	&  --	&	--								\\[.1ex]
Pose Fitting 	&	\multicolumn{5}{c|}{--- 0.29 ---}	   			&  --	&	--								\\ [.1ex]
Expr. Fitting 	&	\multicolumn{5}{c|}{--- 0.30 ---}	   			&  --	&	--								\\[.4ex] \hline
Total			& 0.599	& 16.42	  & 0.90	 & 0.97	& 0.78	&  0.6&	0.088							\\ 
\bottomrule
\end{tabular}
}
\caption{
{\em Expression estimation runtime.} Comparing a number of alternative methods to our ExpNet. Landmark based methods require several steps for landmark detection and then expression optimization; whereas deep methods solve for expression in a single step.
}
\label{tab:timings}
\end{table}

\begin{figure*}[tb]
\centering
\includegraphics[width=\linewidth]{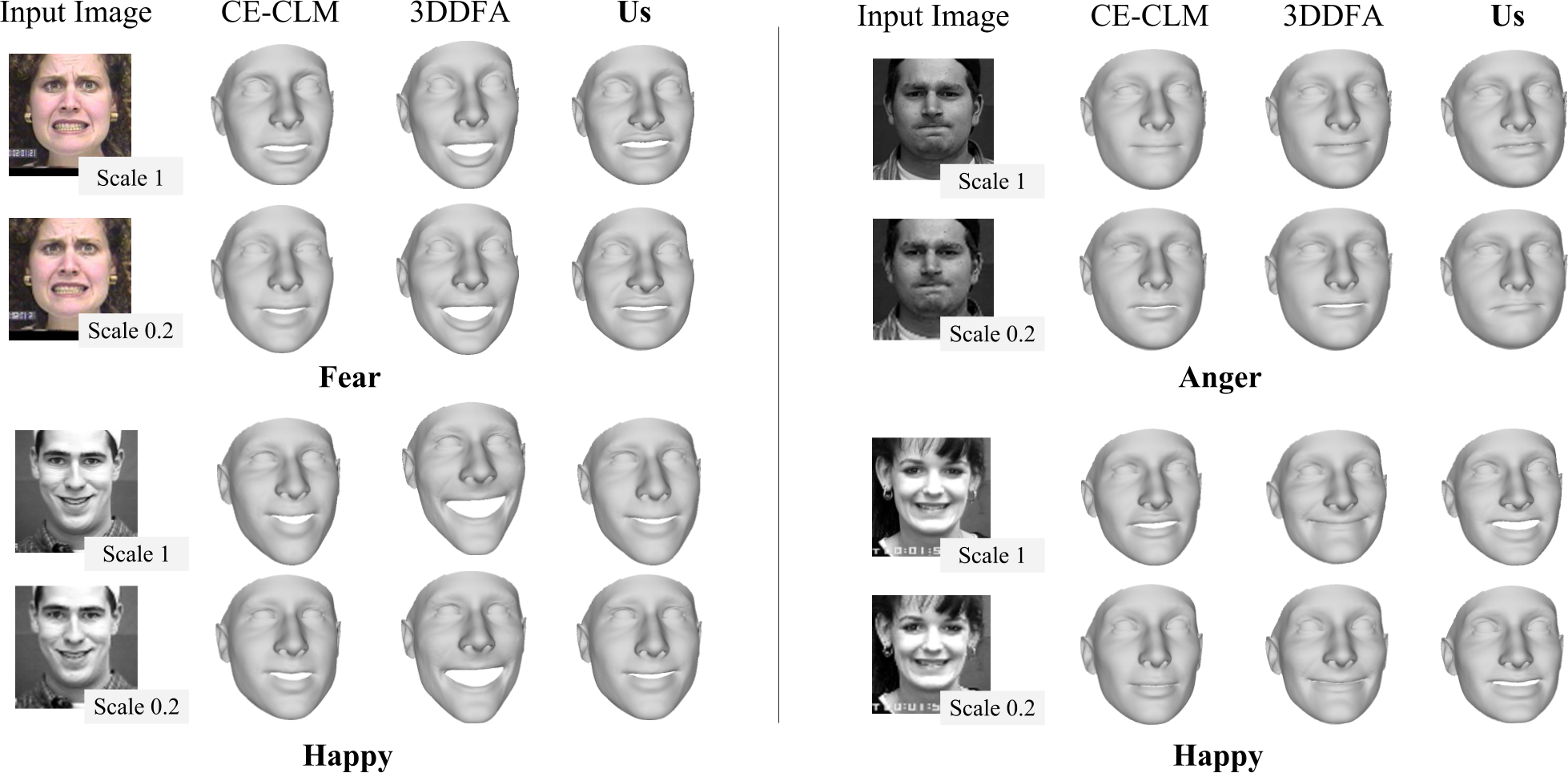}
\caption{
{\em Qualitative expression estimation on CK+}. 3D head shapes estimated by a deep 3DMM fitting method~\cite{tran16_3dmm_cnn}. Expressions added using a number of baseline methods including our ExpNet. Our method is better able to model subtle expressions than 3DDFA. The top-performing landmark detector, CE-CLM~\cite{zadeh2016deep}, does not perform as well on these images.}
\label{fig:qual}
\vspace{-3mm}
\end{figure*}

\begin{figure*}[tb]
\centering
\includegraphics[width=\linewidth]{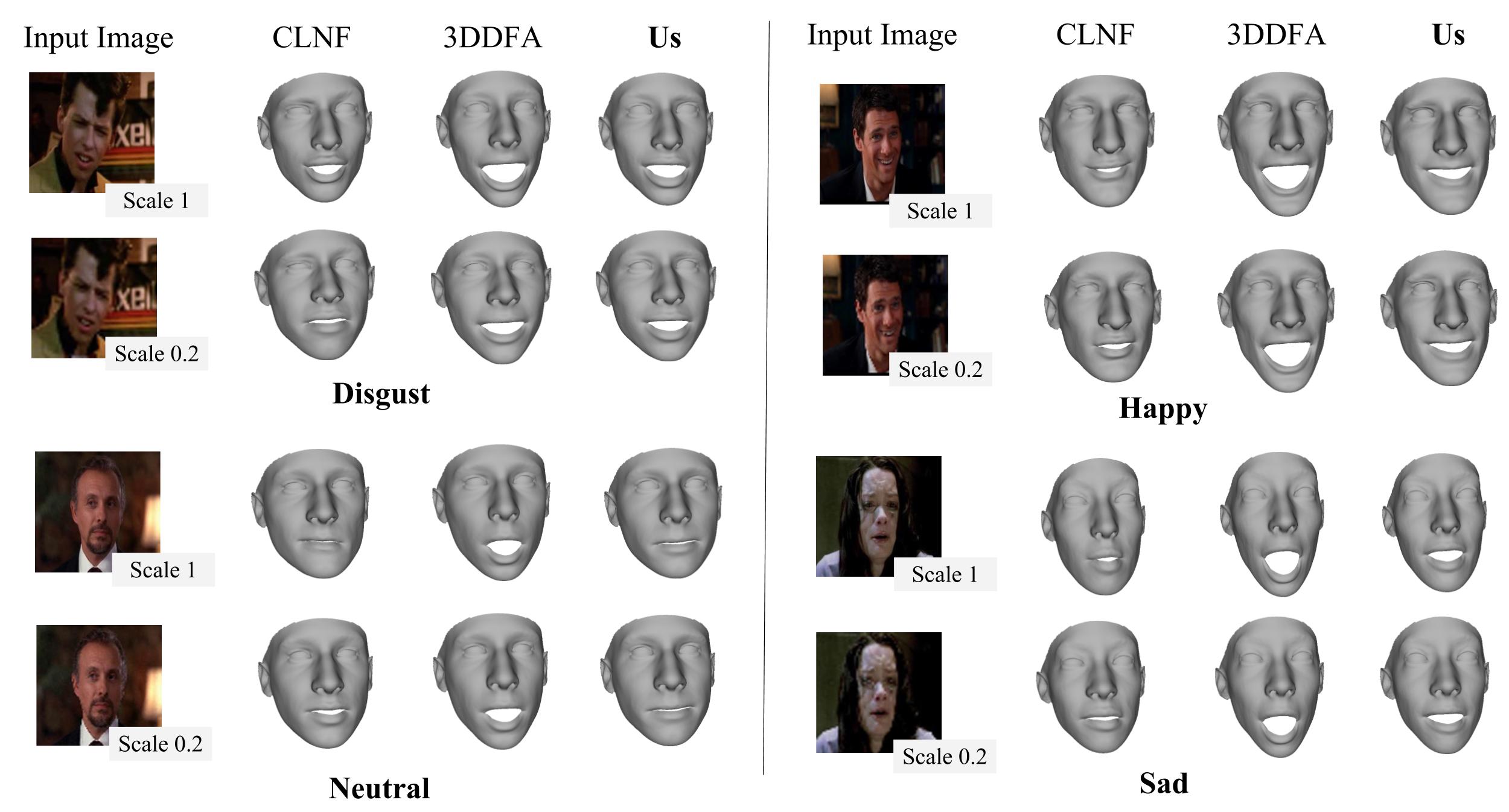}
\caption{
{\em Qualitative expression estimation on EmotiW-17}. 3D head shapes estimated by a deep 3DMM fitting method~\cite{tran16_3dmm_cnn}. We add expressions using a number of baseline methods comparing them with our ExpNet. Our method and 3DDFA~\cite{zhu2015} show consistent expression fitting across scales. Our method additionally models subtle expressions better than 3DDFA. The top-performing facial landmark detector, CLNF~\cite{baltrusaitis2013constrained}, does not perform as well on these images.}
\label{fig:qual_emotiw17}
\vspace{-3mm}
\end{figure*}

\minisection{Baseline methods} We compare our approach to widely used, state-of-the-art face landmark detectors. These are DLIB~\cite{king2009dlib}, CLNF~\cite{baltrusaitis2013constrained}, OpenFace~\cite{baltruvsaitis2016openface}, CE-CLM~\cite{zadeh2016deep}, RCPR~\cite{burgos2013robust}, and 3DDFA~\cite{zhu2015}. Note that CLNF is the method used to produce our training labels. 
 
\minisection{Results} Fig.~\ref{fig:exp_conf} and~\ref{fig:exp_conf_emotiw17} report the emotion classification confusion matrices on the original CK+ and EmotiW-17 datasets (unscaled) for our method (Fig.~\ref{fig:exp_rec_conf_us} and~\ref{fig:exp_rec_conf_us_emotiw17}), comparing it to the other methods, 3DDFA (Fig.~\ref{fig:exp_rec_conf_3ddfa}) and CE-CLM / CLNF (Fig.~\ref{fig:exp_rec_conf_dclm} /~\ref{fig:exp_rec_conf_clnf_emotiw17}). 

On CK+, our expression coefficients were able to capture well surprise (Su), happy (Ha), and disgust (Di) emotions, all emotions which are well defined by facial expressions. On EmotiW-17, our method performed well on neutral (Ne), happy (Ha), sad (Sa), and angry (An), but less so on disgust (Di), fear (Fe), and surprise (Su). From our observations, these last emotions are visually similar to angry (An), which could explain why they challenged our system. On the whole, however, our representation was noticeably better at capturing all emotion classes than its baselines.

Fig.~\ref{fig:exp_rec_scale} reports emotion classification performances of all methods on scaled versions of the CK+ (Fig.~\ref{fig:exp_rec_scale}(a)) and EmotiW-17 sets (Fig.~\ref{fig:exp_rec_scale}(b)). These results measure the sensitivity of different methods to the input image resolution: The x-axis reports the downsizing factor, proportional to the original scale. A scale of~1 therefore represents the original image sizes (640x490 for CK+; 720x576 for EmotiW-17), scale of~0.2 implies 128x98 for CK+ and 144x115 for EmotiW-17, and so fourth. 

Results in Fig.~\ref{fig:exp_rec_scale} clearly show our approach to be the most accurate in terms of emotion recognition accuracy. It is additionally far more robust to scale changes compared than the other landmark detection based methods. Note also the  difference in emotion recognition between deep methods---ours and~\cite{zhu2015}---and landmark based approaches.

Importantly, our method outperforms CLNF~\cite{baltrusaitis2013constrained} by a wide margin in all tests. This result is significant, as CLNF was the method used to generate our expression labels in Sec.~\ref{sec:gen_data}. Our improved performance suggests that the network learned to generalize from its training data and thus performed better on a wider range of viewing conditions and challenges.

\minisection{Runtime} Tab.~\ref{tab:timings} reports runtimes for the methods tested. All tests were performed on a machine with an NVIDIA, GeForce GTX TITAN X and an Intel Xeon CPU E5-2640 v3 @ 2.60GHz. The only exception was 3DDFA~\cite{zhu2015}, which required a Windows system and was tested using an Intel Core i7-4820K CPU @ 3.70GHz with 8 CPUs.

We compare landmark based approaches with deep, direct method such as 3DDFA and our ExpNet. ExpNet is at least one order of magnitude faster than any of its alternatives. Note that, landmark based expression fitting methods generally follow a three-step process: (i) facial landmark detection, (ii) head pose estimation, and (iii) expression fitting. Their total processing time is therefore the sum of these steps. Although some landmark detection methods (e.g. DLIB) are extremely efficient (0.009s), they are still required to solve the optimization problem of Eq.~(\ref{eq:expr_fitting}), in order to translate these detections to an expression coefficients estimate. This process is much slower than our proposed method.

As for deep methods for expression estimation, the software package provided by 3DDFA~\cite{zhu2015} does not allow testing on the GPU; in their paper, they report GPU runtime to be 0.076 seconds, which is similar to our runtime, which was measured on a GPU. Other facial landmarks detector based methods, including the code used to solve~Eq. (\ref{eq:expr_fitting}), are all intrinsically implemented on the CPU. Though they may conceivably be expedited significantly by porting them to the GPU, we are unaware of any such implementation.

\begin{figure}[tb]
\centering
\includegraphics[width=\linewidth]{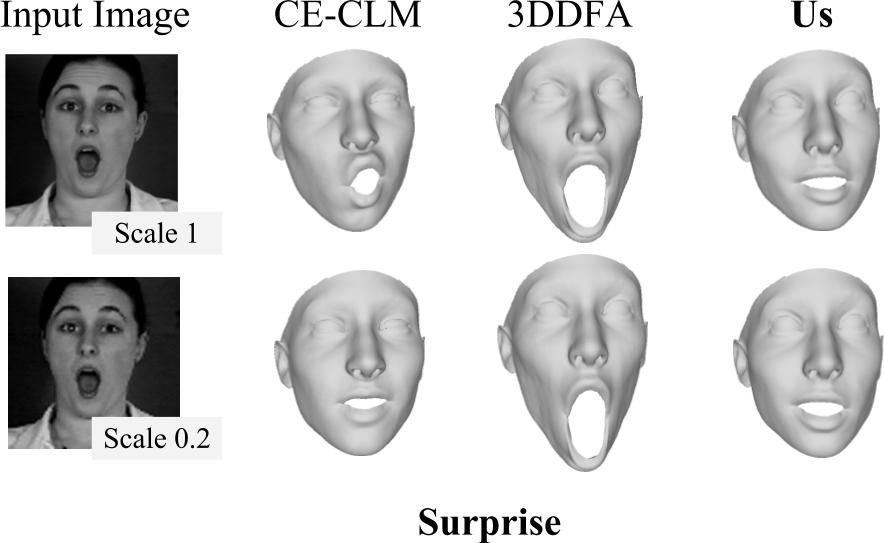}
\caption{
{\em Expression estimation failures.} Our method is less able to handle extreme facial expressions. Other methods, by comparison, appear to either exaggerate the expression (3DDFA) or are inconsistent across scales (CE-CLM).
}
\label{fig:qual_fail}
\end{figure}

\subsection{Qualitative Results}\label{sec:qual}
Fig.~\ref{fig:qual} and~\ref{fig:qual_emotiw17} provide qualitative renderings of the 3D expressions estimated on CK+ and EmotiW-17 images. Each result was obtained on the original, input image scale (scale 1) and also at our lowest resolution (scale 0.2). All the results in these figures use the same 3D shape provided by 3DMM-CNN~\cite{tran16_3dmm_cnn}. Additional, mid level facial details can possibly be added using, e.g.,~\cite{tran2017extreme}, but to emphasize expressions, rather than details, we used only course facial shapes. 

These figures visualize expressions estimated with our ExpNet compared with the recent deep method for joint estimation of shape and expression~\cite{zhu2015}, and the top performing landmark detectors CE-CLM~\cite{zadeh2016deep}, and CLNF~\cite{baltrusaitis2013constrained}. For reference, we provide also the shape 3D face  shape~\cite{tran16_3dmm_cnn} estimated before expressions were added.

Our expression estimates appear to be much better at capturing expression nuances: This is clear from the subtle expressions, fear and anger, rendered in Fig.~\ref{fig:qual}. This is consistent with the improvement shown in the confusion matrices in Fig.\ref{fig:exp_conf}. 3DDFA appears inconsistent across the same expression (happy) and tends to either exaggerate the expression or underestimate it. Both CE-CLM and CLNF seem sensitive to input image resolutions: They both estimate different expressions for the same input image offered at different scales.

Finally, Fig.~\ref{fig:qual} demonstrates a weaknesses of our ExpNet to strong intensity expressions such as surprise. 3DDFA, by comparison, produces somewhat over-exaggerated estimates on these images. Although CE-CLM produces visually suitable estimates, its predictions are inconsistent across scales.

\section{Conclusions}\label{sec:conclu}
We present a method for deep, 3D expression modeling and show it to be far more robust than than facial landmark detection methods widely used for this task. Our approach estimates expressions without the use of facial landmarks, suggesting that facial landmark detection methods may be redundant for this task. This conclusion is consistent with recent results demonstrating deep, {\em landmark free} 3D face shape estimation~\cite{chang17fpn} and 6DoF head alignment~\cite{tran16_3dmm_cnn}. The significance of these results is that by avoiding facial landmark detection, we can process face images obtained in extreme viewing condition which can be challenging for landmark detection methods.
\bibliographystyle{ieee}

\end{document}